# A Qualitative Linear Utility Theory for Spohn's Theory of Epistemic Beliefs

**Phan H. Giang and Prakash P. Shenoy**

University of Kansas School of Business, Summerfield Hall
Lawrence, KS 66045-2003, USA
*pgiang@ukans.edu, pshenoy@ukans.edu*

## Abstract

In this paper, we formulate a qualitative "linear" utility theory for lotteries in which uncertainty is expressed qualitatively using a Spohnian disbelief function. We argue that a rational decision maker facing an uncertain decision problem in which the uncertainty is expressed qualitatively should behave so as to maximize "qualitative expected utility." Our axiomatization of the qualitative utility is similar to the axiomatization developed by von Neumann and Morgenstern for probabilistic lotteries. We compare our results with other recent results in qualitative decision making.

## 1 Introduction

The main goal of this paper is to construct a linear utility theory for lotteries in which uncertainty is described by epistemic beliefs as described by Spohn [20, 21].

Spohn's theory of epistemic beliefs is finding increasing acceptance in artificial intelligence since it is viewed as a qualitative counterpart of Bayesian probability theory. Spohn's theory is also referred to as "kappa calculus." It has its roots in Adams's [1] work on the logic of conditionals, and has been studied extensively by Goldszmidt and Pearl [10, 11] who refer to it as "rank-based system" and "qualitative probabilities." The main representation function in Spohn's calculus is called a disbelief function and its values can be interpreted as infinitesimal or order of magnitude probabilities. Spohn's calculus includes conditional disbelief functions and a notion of conditional independence that satisfies the graphoid axioms [12]. This means that the qualitative theory of (probabilistic) Bayesian networks based on conditional independence applies unchanged to Spohn's calculus. Furthermore the definitions of combination (pointwise addition [18]) and marginalization (minimization) in Spohn's calculus satisfies the axioms described by Shenoy and Shafer [19] that enable local computation. Thus the message-passing architectures for computing marginals such as the Shenoy-Shafer architecture [19] and the Hugin architecture [13] apply also to Spohn's calculus.

One of the major attractions of Bayesian probability theory is a normative decision theory based on von Neumann and Morgenstern's and Savage's theories of rational decision making by maximizing expected utility (or maximizing subjective expected utility in the case of Savage). The focus of this paper is to propose a qualitative linear utility theory for Spohn's calculus so that an analogous decision theory can be formulated for problems in which uncertainty is characterized by epistemic beliefs. We propose axioms analogous to the axioms proposed by von Neumann and Morgenstern (as described by Luce and Raiffa [14]) and describe a representation theorem that states that if the decision makers preferences satisfy these axioms, then there exists a unique qualitative linear utility function such that the utility of any Spohnian lottery is equal to the "expected" utility of the lottery.

An outline of the remainder of this paper is as follows. In Section 2, we briefly describe Spohn's epistemic belief calculus. In Section 3, we define Spohnian lotteries, qualitative utility function, state the axioms, and state and prove the main result. We also describe a small example to illustrate the use of the linear utility function. In Section 4, we discuss the implications of the results and explain the significance of the results using probabilistic semantics of Spohn's calculus. In Section 5, we compare our findings with related research on qualitative decision making theories. Finally in Section 6, we conclude with a summary and some concluding remarks.

## 2 Spohn's Theory of Epistemic Beliefs

Spohn's theory of epistemic beliefs [20, 21, 10] is an elegant, simple and powerful calculus designed to represent and reason with plain human beliefs. The motivation behind Spohn's theory is the need for (i) a formalism to represent plain epistemic beliefs and (ii) procedures for revising beliefs when new information is obtained.

The main ingredients of Spohn's theory are (i) a functional representation of an epistemic state called a disbelief function, and (ii) a rule for revising this function in light of new information. Like a probability distribution function, a disbelief function for a variable is completely specified by its values for the singleton subsets of configurations of the variable.

Formally, let $\Omega$ denote a set of possible worlds. We assume $\Omega$ is finite, $|\Omega| = m$. We use $\omega$ (with subscripts) to denote a world, i.e. $\omega \in \Omega$. If we are interested in a finite set of variables $\{X_1, X_2, \ldots, X_n\}$ each of those is also finite, $\Omega$ can be identified with Cartesian product $\times_1^n \Omega_{X_i}$ where $\Omega_X$ denotes the set of possible values of $X$. Thus, each world $\omega \in \Omega$ is identified with a tuple of values $\langle x_1, x_2, \ldots, x_n \rangle$ where $x_i$ is a value of $X_i$. We also use notation $\omega(i)$ to denote value of variable $X_i$ in the world $\omega$ and $X_i = x$ to denote the subset $\{\omega \in \Omega |\ \omega(i) = x\}$ of $\Omega$.

A Spohnian disbelief function $\delta$ for $\Omega$ is defined as a mapping

$$\delta : 2^\Omega \to Z^+ \cup \{\infty\}$$

where $Z^+$ is set of non-negative integers, satisfying the following axioms:

S1
$$\min_{\omega \in \Omega} \delta(\{\omega\}) = 0$$

and

S2
$$\delta(A) = \begin{cases} \min_{\omega \in A} \delta(\omega) & \text{if } \emptyset \neq A \subseteq \Omega \\ \infty & if A = \emptyset \end{cases}$$

As a result of Axiom (S2), a disbelief function is completely determined by its values for singletons. Thus for computational reasons, we can represent a disbelief function by a disbelief *potential* $\delta : \Omega \to Z^+ \cup \{\infty\}$.

For $A \subseteq \Omega$ such that $\delta(A) < \infty$, the conditional disbelief function $\delta(.|A)$ is defined as

S3 $\quad \delta(B|A) = \delta(B \cap A) - \delta(A).$

It is easy to verify that $\delta(.|A)$ is a disbelief function, i.e., it satisfies $S1$ and $S2$.

The notion of independence for Spohn's epistemic belief is defined similar to that of probability. $A$ and $B$ are independent events if $\delta(A \cap B) = \delta(A) + \delta(B)$. Or in terms of variable we say that $X_i$ and $X_j$ are independent if $\delta(X_i = a, X_j = b) = \delta(X_i = a) + \delta(X_j = b)$ where $a, b$ are arbitrary values of $X_i, X_j$ respectively. It is easy to note that axioms $S1$ through $S3$ which describe the static and dynamic aspects of modeling uncertainty have a similar role to that of Kolmogorov's axioms and Bayes's rule in probability.

To define the semantics of disbelief functions, we will define a related function called a Spohnian *belief* function. Given disbelief function $\delta$, we can define a Spohnian belief function $\beta : 2^\Omega \to Z \cup \{-\infty, \infty\}$ (where $Z$ is the set of all integers) as follows [18]:

$$\beta(A) = \begin{cases} -\delta(A) & \text{if } \delta(A) > 0 \\ \delta(A^c) & \text{if } \delta(A) = 0 \end{cases} \quad (1)$$

where $A^c$ is the complement of $A$ in $\Omega$. $\beta(A) = m$, where $m > 0$, means proposition $A$ is *believed* to degree $m$. $\beta(A) = m$, where $m < 0$, means proposition $A$ is *disbelieved* to degree $m$. $\beta(A) = 0$ means proposition $A$ is neither believed nor disbelieved. $\beta(A) = \infty$ means proposition $A$ is believed with certainty. And $\beta(A) = -\infty$ means proposition $A$ is disbelieved with certainty.

## 3 Spohnian Lotteries and Their Utilities

Following Luce and Raiffa [14], we use the term *Spohnian lottery* to denote a lottery in which uncertainty is modeled by a Spohnian disbelief function. Let $\mathcal{O} = \{o_1, o_2, \ldots, o_r\}$ denote a finite set of prizes involved in a lottery. We assume, without loss of generality, a strict preference order over the set of prizes $o_1 \succ o_2 \succ \ldots \succ o_r$ where $\succ$ reads "is qualitatively preferred to". So there are no two equally preferred prizes in $\mathcal{O}$. $o_1$ is the best prize, and $o_r$ is the worst.

A *simple* Spohnian lottery is a pair of a prize vector and a vector representing a Spohnian disbelief potential. We write it in the form $(O.\delta)$ where $O$ is the prize vector $(o_1, o_2, \ldots, o_r)$ and $\delta = (\delta_1, \delta_2, \ldots, \delta_r)$ is a Spohnian disbelief potential vector, i.e., $\delta_i \in Z^+ \cup \{\infty\}$ and

$$\min_{1 \leq i \leq r} \delta_i = 0.$$

Sometimes $[o_1.\delta_1, o_2.\delta_2, \ldots, o_r.\delta_r]$ is used to denote a lottery. And because the set of prizes is fixed, a simple lottery can be identified with just a disbelief vector $(\delta_1, \delta_2, \ldots, \delta_r)$. We use the convention that a prize is a simple lottery in which the disbelief degree associated with the prize is zero and others are infinity. This leads us to the concept of *compound* or *multi-stage* lotteries in which the prizes are again lotteries. We use **L** to denote the set of all lotteries, simple or compound.

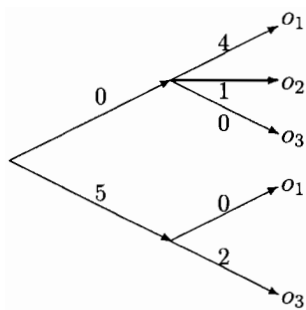

Figure 1: A lottery tree of depth 2.

Graphically, a lottery is a rooted tree whose leaves are prizes and associated with branches outgoing from a node is a (conditional) disbelief function. So we can define a lottery's depth as the depth of the corresponding tree. For example, a prize is a lottery of depth 0, a simple lottery has depth 1, and so on. In Figure 1, a graphical representation of a lottery of depth 2, $[[o_1.4, o_2.0, o_3.0].0, [o_1.0, o_3.2].5]$, is shown.

A *standard* lottery is one that realizes in either the most preferred prize $o_1$ or the least preferred prize $o_r$, i.e., $[o_1.\kappa_1, o_r.\kappa_r]$. Of course, condition $\min(\kappa_1, \kappa_2) = 0$ must also be satisfied. We use **S** to denote the set of all standard lotteries.

Since our task is to compare lotteries, we will study a complete and transitive relation[1] $\succeq$ on the set **L**. Notice that $\succeq$ denotes a qualitative preference relation on **L**, i.e., $L_1 \succeq L_2$ means $L_1$ is "approximately preferred or indifferent to" $L_2$. This preference relation will be represented by what we call a "qualitative utility function." To enable this representation, we assume that the preference relation satisfies some desirable properties. Formally, we will adopt axioms similar to those presented in Luce and Raiffa [14].

**Axiom 1 (Ordering of prizes)** *The preference relation $\succ$ over the set of prizes $\mathcal{O}$ is complete and transitive.*

This axiom simply formalizes our assumption about the set of prizes.

**Axiom 2 (Reduction of compound lotteries)**
*Any compound lottery is indifferent to a simple lottery whose disbelief degrees are calculated according to Spohn's calculus. A compound lot-*

---
[1]Derivative relations $\succ$ and $\sim$ are defined from $\succeq$ as usual, i.e., $a \succ b$ means $a \succeq b$ and $b \not\succeq a$, and $a \sim b$ means $a \succeq b$ and $b \succeq a$.

*tery $L_c = [L_1.\delta_1, L_2.\delta_2, \ldots, L_k.\delta_k]$ where $L_i = [o_1.\kappa_{i1}, o_2.\kappa_{i2}, \ldots, o_r.\kappa_{ir}]$ for $1 \leq i \leq k$ is indifferent to the simple lottery $L_s = [o_1.\kappa_1, o_2.\kappa_2, \ldots, o_r.\kappa_r]$ where*

$$\kappa_j = \min_{1 \leq i \leq k} \{\delta_j + \kappa_{ij}\} \quad (2)$$

The intuition behind this axiom is as follows. The compound lottery $L_c$ can be interpreted as two stage process. The outcomes possible for the first stage are $x_1, x_2, \ldots, x_k$. If $x_i$ realizes, the lottery player gets simple lottery $L_i$ which in turn has set of outcomes $y_1, y_2, \ldots, y_r$. If $y_j$ realizes, the player is rewarded with $o_j$. Another way to view $L_c$ is by collapsing the two stages together to obtain a lottery with the set of possible outcomes $\{\langle x_iy_j\rangle | 1 \leq i \leq k \text{ and } 1 \leq j \leq r\}$. When a $\langle *y_j \rangle$ is realizes (wildcard $*$ can match with any $x$), the player is rewarded with prize $o_j$. So the degree of disbelief the player associates with getting prize $o_j$ is disbelief degree he assigns to set $*y_j$. As per Spohn's calculus, the disbelief degree associated with combined state $x_iy_j$ is calculated by $\delta_i + \kappa_{ij}$. Hence, disbelief degree assigned to set $*y_j$ is $\min\{\delta_j + \kappa_{ij} | 1 \leq i \leq k\}$.

**Axiom 3 (Substitutability)**
*Indifferent lotteries are substitutable. That is if $L_i \sim L_i'$ then $[L_1.\delta_1, \ldots, L_i.\delta_i \ldots, L_k.\delta_k] \sim [L_1.\delta_1, \ldots, L_i'.\delta_i, \ldots, L_k.\delta_k]$.*

This requirement is the same as presented in Luce and Raiffa and it conveys the idea that preference relation reflects the desirability of a lottery and that desirability is not context sensitive.

**Axiom 4 (Quasi-continuity)** *For each prize $o_i$, there exists a standard lottery that is indifferent to it.*

In particular, we assume $o_1 \sim [o_1.0, o_r.\infty]$ and $o_r \sim [o_1.\infty, o_r.0]$. This is reasonable since in $[o_1.0, o_r.\infty]$, we believe in $o_1$ with certainty, and in $[o_1.\infty, o_r.0]$, we believe in $o_r$ with certainty. A comparison with the continuity assumption adopted for probabilistic case may give an impression that this assumption is too strong because set of standard (qualitative) lotteries does not constitute a continuum. We hold that it is quite reasonable since $\succeq$ should be read as "qualitatively preferred to". One can regard the set of standard lotteries as a fishing net that spreads from most preferred lottery $o_1$ to the least preferred one $o_r$. Therefore, any lottery $L$ is caught between a pair of successive knots, for example, between say $[o_1.0, o_r.k]$ and $[o_1.0, o_r.k+1]$. So what this axiom entails is to disallow the ambivalence and force a "qualitative indifference" between $L$ and one of the two standard lotteries.

**Axiom 5 (Transitivity)** *Preference relation over the set of lotteries is complete and transitive. For-*

mally, for $L_i, L_j, L_k \in \mathbf{L}$, either $L_i \succeq L_j$ or $L_j \succeq L_i$; and if $L_i \succeq L_j$ and $L_j \succeq L_k$, then $L_i \succeq L_k$

**Axiom 6 (Qualitative monotonicity)** *The preference relation $\succeq$ over $\mathbf{S}$ satisfies the following condition. Suppose $s = [o_1.\kappa_1, o_r.\kappa_r]$ and $s' = [o_1.\kappa'_1, o_r.\kappa'_r]$ are two standard lotteries then*

$$s \succ s' \quad \textit{iff} \quad \begin{cases} \kappa_1 = \kappa'_1 = 0 \ \& \ \kappa_r > \kappa'_r & (i) \\ \kappa_1 = 0 \ \& \ \kappa'_1 > 0 & (ii) \\ \kappa_1 < \kappa'_1 \ \& \ \kappa_2 = \kappa'_2 = 0 & (iii) \end{cases} \quad (3)$$

The intuition behind this axiom is as follows. In the first case $(i)$ when $\kappa_1 = \kappa'_1 = 0$, we believe (using function $\beta$ given in equation (1)) $o_1$ to degree $\kappa_r$ in $s$ and we believe $o_1$ to degree $\kappa'_r$ in $s'$. Since $\kappa_r > \kappa'_r$ one should prefer $s$ to $s'$. In the second case, we believe $o_1$ (to degree $\kappa_r$) in $s$ and we believe in $o_r$ (to degree $\kappa'_1$) in $s'$. Again, one should prefer $s$ to $s'$ (regardless of the values of $\kappa_r$ and $\kappa'_1$). In the third case $(iii)$, we believe $o_r$ to degree $\kappa_1$ in $s$, and we believe $o_r$ to degree $\kappa'_1$ in $s'$. Since $\kappa_1 < \kappa'_1$, one should prefer $s$ to $s'$. Thus, on the set of standard lotteries $\mathbf{S}$ there is a well defined (complete and transitive) preference relation.

We have the following lemma that states that the set of Spohnian lotteries divided by the indifference relation is isomorphic to the set of standard lotteries.

**Lemma 1** *If the preference relation $\succeq$ on the set of lotteries $\mathbf{L}$ satisfies axioms 1 though 6, then for each lottery there exists one and only one standard lottery indifferent to it.*

**Proof:** We prove the existence of indifferent standard lottery by induction on the depth of lottery trees.

For a constant lottery (of depth 0), because of axiom 4, each prize $o_i$ is indifferent to a standard lottery $s_i$.

A lottery of depth 1 is either a standard lottery or a simple lottery. Obviously, a standard lottery is indifferent to itself. For a simple lottery $L = [o_1.\delta_1, o_2.\delta_2, \ldots, o_r.\delta_r]$, by axiom 3, $L \sim [s_1.\delta_1, s_2.\delta_2, \ldots, s_r.\delta_r]$. By axiom 2, the latter can be reduced to a standard lottery $s$ such that $[s_1.\delta_1, s_2.\delta_2, \ldots, s_r.\delta_r] \sim s$.

Suppose for all lotteries of depth not greater than $n$, there is a standard lottery indifferent to it.

For a lottery $L$ of depth $n + 1$. This lottery is a compound lottery whose prizes are lotteries of depth not greater than $n$. Because of induction hypothesis, each prize of $L$ is indifferent to a standard lottery. By substitutability, $L$ is indifferent to a compound lottery of depth 2. Again by induction hypothesis, there is a standard lottery indifferent to it.

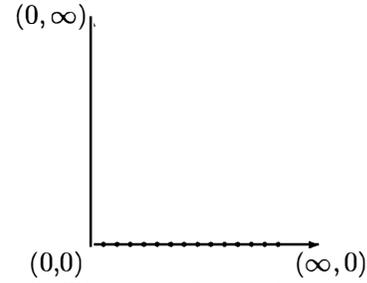

Figure 2: The space $\mathbf{B}_0$ depicted by dots.

Finally, we have to show that there is only one standard lottery indifferent to a given lottery. Suppose there are two standard lotteries $s_1, s_2 \in \mathbf{S}$ such that $s_1 \sim L$ and $s_2 \sim L$. By axiom 5, we have $s_1 \sim s_2$. But by axiom 6, it is possible only if $s_1 = s_2$. ∎

From a decision theoretic perspective, we would like to model a preference relation $\succeq$ on the set of all lotteries $\mathbf{L}$ by a utility function $u : \mathbf{L} \to Z \cup \{-\infty, \infty\}$ such that given any two lotteries $L$ and $L'$, $u(L) \geq u(L')$ if and only if $L \succeq L'$. Clearly, this implies that $u(L) = u(L')$ whenever $L \sim L'$. Notice that unlike traditional quantitative utility function which has range in the real line, the function $u$ has value in a discrete set which in this case is the set of integers and the labels $-\infty$ and $\infty$. From Lemma 1, it is clear that if we find a way to assign utility values to standard lotteries then it is straightforward to do so for any lottery.

Next we define an utility function for standard lotteries. We abbreviate $s = [o_1.\kappa_1, o_r.\kappa_r]$ by a pair $(\kappa_1, \kappa_r)$. From the qualitative monotonicity axiom, it is clear that the following function will satisfy the definition of a qualitative utility function above:

$$u_s((\kappa_1, \kappa_r)) = \kappa_r - \kappa_1 \quad (4)$$

For maintaining the analogy with the case of probabilistic lotteries, we will define a utility function as a function $U : \mathbf{L} \to \mathbf{B}_0$ where $\mathbf{B}_0$ is defined as follows:

$$\mathbf{B}_0 \stackrel{\text{def}}{=}$$
$$\{(x,y)|x, y \in Z^+ \cup \{\infty\} \text{ s.t. } \min(x,y) = 0\}$$

Even though $\mathbf{B}_0$ is a subset of $(Z^+ \cup \{\infty\}) \times (Z^+ \cup \{\infty\})$, we can define a complete and transitive order $\geq$ on the set $\mathbf{B}_0$ as follows: $(x_1, y_1) \geq (x_2, y_2)$ if and only if $y_1 - x_1 \geq y_2 - x_2$. Alternatively, notice that Equation 4 establishes an isomorphism between $\mathbf{B}_0$ and $Z \cup \{-\infty, \infty\}$. Therefore, $\mathbf{B}_0$ inherits all the order relations of $Z \cup \{-\infty, \infty\}$. Thus a function $U : \mathbf{L} \to \mathbf{B}_0$ is a *qualitative utility function* if $U(L) \geq U(L')$ iff $L \succeq L'$.

Since $\mathbf{B}_0$ is a set of binary vectors, addition of a scalar, and pointwise minimization are defined as usual. Suppose $c \in Z^+$, $\mathbf{b} = (\mathbf{x}, \mathbf{y})$ and $\mathbf{b_i} = (\mathbf{x_i}, \mathbf{y_i})$

$$c + \mathbf{b} \stackrel{\text{def}}{=} (x + c, y + c) \quad (5)$$

$$\min_i \{\mathbf{b_i}\} \stackrel{\text{def}}{=} (\min_i \{x_i\}, \min_i \{y_i\}) \qquad (6)$$

Next, we state and prove a "qualitative linear utility" representation theorem that is analogous to the representation theorem of von Neumann and Morgenstern.

**Theorem 1** *Suppose we are given a preference relation $\succeq$ on the set of all lotteries $\mathbf{L}$ that satisfies Axioms 1 to 6. There exists a qualitative utility function $U : \mathbf{L} \to \mathbf{B}_0$ such that*

$$U([L_1.\delta_1, L_2.\delta_2, \ldots, L_k.\delta_k]) = \min_{1 \leq i \leq k} \{\delta_i + U(L_i)\} \qquad (7)$$

*Futhermore, such a qualitative utility function is unique.*

**Proof:** First we prove the existence of a qualitative utility function $U : \mathbf{L} \to \mathbf{B}_0$ by constructing it as follows. For standard lotteries, $U$ is defined as follows:

$$U([o_1.\kappa_1, o_r.\kappa_r]) \stackrel{\text{def}}{=} (\kappa_1, \kappa_r) \qquad (8)$$

For an arbitrary lottery $L$, we define $U(L) = U(s)$ where $s$ is the standard lottery that is indifferent to $L$. By Lemma 1, each lottery $L$ is indifferent to exactly one standard lottery $s$. Therefore the function $U$ is a well-defined qualitative utility function.

Next, we will show that U as constructed above satisfies Equation 7. Consider depth-one lottery $L = [o_1.\delta_1, o_2.\delta_2, \ldots, o_r.\delta_r]$. By Axiom 4, each prize $o_i$ is indifferent to a standard lottery, say $s_i = [o_1.\kappa_{i1}, o_r.\kappa_{ir}]$. Therefore, $U(o_i) = (\kappa_{i1}, \kappa_{ir})$. Consider lottery $L' = [s_1.\delta_1, s_2.\delta_2, \ldots, s_r.\delta_r]$. From Axiom 3, $L \sim L'$. By Axiom 2, $L'$ is indifferent to the standard lottery $s = [o_1.\kappa_1, o_r.\kappa_r]$ where

$$\kappa_1 = \min_{1 \leq i \leq r} \{\delta_i + \kappa_{i1}\} \text{ and } \kappa_r = \min_{1 \leq i \leq r} \{\delta_i + \kappa_{ir}\} \qquad (9)$$

Therefore $U(L) = U(L') = U(s) = (\kappa_1, \kappa_r)$. Finally we notice that

$$\begin{aligned}
&\min_{1 \leq i \leq r} \{\delta_i + U(o_i)\} \\
&= \min_{1 \leq i \leq r} \{\delta_i + (\kappa_{i1}, \kappa_{ir})\} \\
&= \min_{1 \leq i \leq r} \{(\delta_i + \kappa_{i1}, \delta_i + \kappa_{ir})\} \\
&= (\min_{1 \leq i \leq r} \{\delta_i + \kappa_{i1}\}, \min_{1 \leq i \leq r} \{\delta_i + \kappa_{ir}\}) \\
&= (\kappa_1, \kappa_r)
\end{aligned}$$

Therefore $U(L) = \min_{1 \leq i \leq r} \{\delta_i + U(o_i)\}$. By induction on the lottery's depth, we can prove this property for any general lottery.

The proof of the fact that $U$ defined above is the only qualitative utility function satisfying Equation 7 breaks down into several small steps. Let $u$ be another qualitative utility function from $\mathbf{L}$ to $\mathbf{B}_0$ satisfying Equation 7.

First, we will show that $u$ has value in both "half lines" $\{(0, y)\}$ and $\{(x, 0)\}$. Suppose to the contrary, $u(L) \geq (0, h)$ for all $L \in \mathbf{L}$. Let us denote standard lottery $[o_1.a, o_r.b]$ by $s_{\{a,b\}}$. Let $u(s_{\{0,0\}}) = (0, k)$ and $u(s_{\{m,0\}}) = (0, k')$. Because $s_{\{0,0\}} \succ s_{\{m,0\}}$, and $u$ is utility function, we have $k > k'$. Consider lottery $[s_{\{0,0\}}.0, s_{\{m,0\}}.0]$. We have $[s_{\{0,0\}}.0, s_{\{m,0\}}.0] \sim s_{\{0,0\}}$. Since $u$ is a qualitative utility function $u([s_{\{0,0\}}.0, s_{\{m,0\}}.0]) = u(s_{\{0,0\}}) = (0, k)$. On the other hand, applying the formula in the right-hand side of Equation 7, we have $u([s_{\{0,0\}}.0, s_{\{m,0\}}.0]) = \min\{(0, k), (0, k')\} = (0, k')$ leading to a contradiction. Therefore $u$ has values in both half lines of $\mathbf{B}_0$.

We will now show that $u(s_{\{0,0\}}) = (0, 0)$. Suppose to the contrary $u(s_{\{0,0\}}) = (0, k)$ with $k > 0$. From the previous step, we can assume there are lotteries $s_1, s_2$ such that $u(s_1) = (0, k)$ and $u(s_2) = (h, 0)$. By considering lotteries of the form $[s_1.0, s_2.\delta]$ or $[s_1.\delta, s_2.0]$, using the right-hand side of Equation 7, we see that points $(0, \delta)$ and $(\delta, 0)$ with $\delta \leq \min\{k, h\}$ represent values of $u$ for some lotteries. In particular there is a standard lottery $s$ such that $u(s) = (0, 0)$. Since $u(s_{\{0,0\}}) = (0, k) > (0, 0) = u(s)$, $s$ must have the form $s_{\{m,0\}}$ for some $m > 0$ i.e. $u(s_{\{m,0\}}) = (0, 0)$. Consider again the lottery $[s_{\{0,0\}}.0, s_{\{m,0\}}.0]$. We have $[s_{\{0,0\}}.0, s_{\{m,0\}}.0] \sim s_{\{0,0\}}$. So, $u([s_{\{0,0\}}.0, s_{\{m,0\}}.0]) = u(s_{\{0,0\}}) = (0, k)$. On the other hand, applying the formula in the right-hand side of Equation 7, we have $u([s_{\{0,0\}}.0, s_{\{m,0\}}.0]) = \min\{(0, k), (0, 0)\} = (0, 0)$ leading to a contradiction. Therefore $u(s_{\{0,0\}}) = (0, 0)$.

We can use the Dirichlet principle to show that $u(s_{\{0,m\}}) = (0, m)$. Suppose $u(s_{\{0,m\}}) = (0, n)$. Consider case $n < m$. Since $s_{\{0,m\}} \succ s_{\{0,m-1\}} \succ \ldots \succ s_{\{0,0\}}$, we have $m + 1$ different lotteries but only $n + 1$ slots for utility values from $(0, 0)$ to $(0, n)$. Thus, two lotteries one of which is strictly preferred to the other, must be given the same value by $u$ leading to a contradiction. In case $n > m$, consider lotteries of the form $[s_{\{0,0\}}.\delta, s_{\{0,m\}}.0]$ where $\delta \leq n$. Applying the right-hand side of Equation 7 for them, we see that every of $n + 1$ slots from $(0, 0)$ to $(0, n)$ must be filled by a lottery in $\mathbf{L}$. Using Lemma 1 these slots must be filled by standard lotteries. But between $s_{\{0,0\}}$ to $s_{\{0,m\}}$, there are only $m + 1$ standard lotteries again leading to a contradiction.

Thus, if $u$ is a qualitative utility function satisfying equation 7 then $u = U$. In other words, $U$ is unique. ∎

Figure 3 illustrates the calculation of expected utility of the lottery in Figure 1 if we assume $o_2 \sim [o_1.0, o_3.3]$ and $o_1$ is the best prize and $o_3$ is the worst prize.

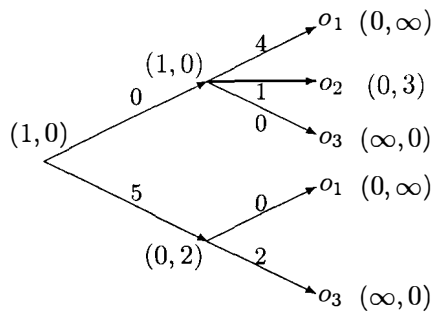

Figure 3: Expected utility of a lottery.

Notice that when $L$ is a simple lottery, Equation 7 can be rewritten as

$$U([o_1.\delta_1, o_2.\delta_2, \ldots, o_r.\delta_r]) = \min_{1 \leq i \leq r} \{\delta_i + U(o_i)\}$$

which is structurally similar to von Neumann - Morgenstern's expected utility formula. Multiplication in probability is replaced by addition in Spohn's theory. And addition in probability theory is replaced by minimization in Spohn's theory. Therefore, we refer to the right hand side of Equation 7 as "qualitative expected utility." We have used $\mathbf{B}_0$ as a scale for measuring preferences. However, this is done only to show that the form of the utility function in Equation 7 is analogous to the von Neumann-Morgenstern utility for probabilistic lotteries. Using Equation 4, we can use $Z \cup \{-\infty, \infty\}$ as the scale of the utility function.

One difference between qualitative and quantitative lotteries should be noted. While quantitative utility uses a continuous scale (real numbers), qualitative utility use a discrete scale. So one should not expect a smooth gradation in utility for qualitative lotteries as is the case for probabilistic lotteries. For example, consider two prizes $o_i \succ o_{i+1}$ that are successive in the sense that there is no other prize in between them i.e. there is no $o_j$ such that $o_i \succ o_j \succ o_{i+1}$. Consider a lottery of the form $[o_i.\delta_i, o_{i+1}.\delta_{i+1}]$ where $\min(\delta_i, \delta_{i+1}) = 0$ and $\delta_i, \delta_{i+1} < \infty$. Intuition from quantitative utility would suggest $U(o_i) > U([o_i.\delta_i, o_{i+1}.\delta_{i+1}]) > U(o_{i+1})$. However, the following example shows that this relationship does not always hold in the case of qualitative utility. Let us assume $U(o_i) = (0, \kappa+\sigma)$ and $U(o_{i+1}) = (0, \kappa)$. In the case $\delta_{i+1} > \delta_i = 0$, by Theorem 1, $U([o_i.0, o_{i+1}.\delta_{i+1}]) = \min\{U(o_i), \delta_{i+1} + U(o_{i+1})\} = (0, \min\{\kappa + \sigma, \kappa + \delta_{i+1}\}$. If we further assume that $\sigma \leq \delta_{i+1}$ then $U([o_i.0, o_{i+1}.\delta_{i+1}]) = (0, \kappa+\sigma) = U(o_i)$. Thus the qualitative utility scale is unable to always make fine distinctions that the probabilistic utility theory can make.

**Example:** Building Houses in an Earthquake Zone.

**Outcomes.** Houses that can survive an earthquake of intensity k where $0 \leq k \leq 12$ are measured in Mercalli Intensity Scale[2], which ranks earthquakes in terms of magnitude of destruction they cause for structures: $q_0, q_1, \ldots, q_{12}$.

**Qualitative Utility of an Earthquake-Proof Residence.** Let us consider the following hypothetical situation. A person is planning to build her house in a earthquake-prone region. When the homeowner considers what earthquake intensity her house should withstand, she may not be able to relate a certain intensity, say $q_5$, with an uncertain situation involving only no earthquake ($q_0$) and the most intense earthquake ($q_{12}$). However, in terms of financial cost, obviously the higher the earthquake intensity a house can survive the costlier it is. So, the homeowner will definitely prefer $q_0 \succ q_1 \succ \ldots \succ q_{12}$. We can associate an earthquake of intensity $k$ with a pair of numbers $(a_k, b_k)$ as follows. The damage caused by earthquakes can range from *no perceptible damage* to *complete destruction*. Thus we set $o_1$ as no damage and $o_r$ as complete destruction. For each earthquake intensity, we find an equally preferred standard lottery and denote the disbelief function by $(a_k, b_k)$. An example of one set of such assessments is shown in Table below. One interpretation of the assessments is that the damage caused by earthquakes of intensity up to 2 is viewed as "acceptable" since they are all equally preferred to standard lotteries where the homeowner believes the outcome is no damage (to different degrees, of course), and damage caused by earthquakes of intensity 4 or more are "unacceptable" since they are all equally preferred to standard lotteries where the homeowner believes the outcome is complete destruction (to different degrees). The damage caused by intensity 3 earthquake lies between these two categories.

Given that "scientists have never predicted a major earthquake, nor do they know how or expect to know how any time in the foreseeable future" (The United States Geological Surveys' web site), it seems there is not sufficient information to produce a probability distribution of earthquakes in a certain region. However, the subjective epistemic belief about EQ occurrence may be represented in the form of a Spohnian epistemic disbelief function $\delta_0, \delta_1, \ldots, \delta_{12}$, i.e., the homeowner believes (to degree 1) that an earthquake of intensity 4 will occur during her ownership of her new home.

---

[2]Level 0 represents no earthquake. A level 1-5 on the Mercalli scale would represent a small amount of observable damage. For example, at this level doors would rattle, dishes break and weak or poor plaster would crack. As the level rises toward the larger numbers, the amount of damage increases considerably. The number 12 represents total damage. List of 1-12 levels of the Modified Mercalli Intensity Scale of 1931 (Abridged; Wood and Neumann, 1931). See Earthquakes & Volcanoes, v. 25, no. 2, 1994, p. 87 for more details.

The above situation can be seen to satisfy the axioms of qualitative utility. Therefore, we can estimate "expected utility" of the situation.

$$U([q_0.\delta_0, \ldots, q_{12}.\delta_{12}]) = \min_{0 \leq i \leq 12} \{\delta_i + U(q_i)\} \quad (10)$$

Suppose the information is given in the following table

| EQ Intensity | $q_0$ | $q_1$ | $q_2$ | $q_3$ | $q_4$ | $q_5$ | $q_6$ |
|---|---|---|---|---|---|---|---|
| Utility | 0 | 0 | 0 | 0 | 2 | 3 | 4 |
| assessment | $\infty$ | 7 | 2 | 0 | 0 | 0 | 0 |
| $\delta_i$ | 4 | 3 | 2 | 1 | 0 | 1 | 2 |

| $q_7$ | $q_8$ | $q_9$ | $q_{10}$ | $q_{11}$ | $q_{12}$ |
|---|---|---|---|---|---|
| 6 | 9 | 11 | 14 | 18 | 21 |
| 0 | 0 | 0 | 0 | 0 | 0 |
| 2 | 3 | 4 | 5 | 6 | 7 |

The calculation of qualitative expected utility of the situation results in $(1,0)$. That ranks the given uncertain situation in between earthquakes of intensity 3 and 4, i.e., slightly unacceptable. In other words, the prospective homeowner should build her home so that it can survive an earthquake of intensity 4. ∎

## 4 Qualitative vs. Quantitative Utility

So far, we have discussed essentially ordinal relationship among qualitative lotteries. Thanks to the ordinal semantics, the proposed qualitative utility theory is well suited for situations when quantitative assessment of the strength of belief and/or the desirability of consequences is difficult. Various reasons contribute to that difficulty such as the nature of a problem, and the subjective ability of assessors, or the cost of doing an assessment. In practice, ordinal information is often represented by numbers. Thus in practice, information is rarely purely quantitative or purely qualitative, but somewhere in between. For example, by assuming an objective probability distribution for prizes of lotteries, utility theory in probabilistic lottery framework considered by von Neumann and Morgenstern is quantitative. But when the theory is applied in practice, often, the required probability distribution is obtained through some conversion of subjective opinions or sparse statistical data. In that sense, the application of the theory is somewhat qualitative.

For the qualitative utility theory based on Spohn epistemic belief developed here, the question we address in this section is how can quantitative information be used when it is available. It is well known that one interpretation of Spohnian disbelief degree is the order-of-magnitude approximation of probabilities [20, 21, 10, 5, 23, 11]. The idea is to express probability as a polynomial function of some $\epsilon > 1$

$$P_\epsilon(\omega) = \sum_{i \geq 0}^{n} a_i * \epsilon^{-i} \quad (11)$$

where $0 \leq a_i < \epsilon$. That is, $\overline{0 \cdot a_0 a_1 \ldots a_n}$ is a numerical representation of $P(\omega)$ in the $\epsilon$-base system. Then the degree of disbelief is the absolute value of the order of the polynomial which is the smallest index with strictly positive coefficient. Note that since $P_\epsilon(\omega) \leq 1$ and $\epsilon > 1$, the order of $P_\epsilon$ is non-positive. The same result can be obtained though a logarithmic transformation $\kappa(\omega) = \lfloor -\log_\epsilon(P(\omega)) \rfloor$. Suppose $\epsilon = 10$, for example, $.325 = 3 * \epsilon^{-1} + 2 * \epsilon^{-2} + 5 * \epsilon^{-3}$. Thus, the degree of disbelief associated with a probability in the interval $[.1, 1]$ is 0. When probability is in $[.01, .1)$ disbelief degree is 1, and so on. In other words, we have the following rule: the degree of disbelief is the number of leading zeros in the $\epsilon$-based representation of a probability.

Thus, Spohn's calculus can be interpreted in the light of manipulation of orders of probability polynomials. The order of sum of two polynomials equals the maximum of the two orders. And the order of product of two polynomials is the sum of the two orders.

The idea of representing a number as a polynomial function can also shed some light on the expected qualitative utility formula (7). Let us consider a von Neumann-Morgenstern lottery $L = [p_1.o_1, p_2.o_2, \ldots, p_n.o_n]$ where $p_i$ is the probability of winning prize $o_i$. Assume $o_1 \succ o_2 \succ \ldots o_n$ and also that utility of prizes is normalized i.e. $u(o_1) = 1$ and $u(o_n) = 0$. The expected utility of the lottery $u(L)$ is shown to be

$$u([p_1.o_1, p_2.o_2, \ldots, p_n.o_n]) = \sum_{i=1}^{n} p_i * u(o_i) \quad (12)$$

Now let us express $p_i$, $u(L)$ and $u(o_i)$ as polynomials of some $\epsilon > 1$ i.e. $p_i = P_i(\epsilon)$, $u(L) = L(\epsilon)$ and $u(o_i) = O_i(\epsilon)$. We shall abuse notation slightly by considering $\kappa$ as operator that extracts the absolute value of the order of a polynomial i.e. $\kappa(P(\epsilon))$ is the absolute value of order of $P(\epsilon)$. Now applying $\kappa$ operator on both sides of equation (12) with value replaced by corresponding polynomials we have

$$\kappa(L(\epsilon)) = \kappa(\sum_{i=1}^{n} P_i(\epsilon) * O_i(\epsilon)) \quad (13)$$

By definition of $\kappa$, the right hand side of (13) expands

to

$$\kappa(\sum_{i=1}^{n} P_i(\epsilon) * O_i(\epsilon)) = \min_{i}\{\kappa(P_i(\epsilon)) + \kappa(O_i(\epsilon))\}. \quad (14)$$

So,

$$\kappa(L(\epsilon)) = \min_{i}\{\kappa(P_i(\epsilon)) + \kappa(O_i(\epsilon))\}. \quad (15)$$

Comparing (15) and (7), we see that the expected qualitative utility theorem is in agreement with (quantitative) expected utility theorem if we interpret disbelief degree and qualitative utility as order-of-magnitude abstraction of probability and von Neumann-Morgenstern utility respectively.

## 5　Related Work

In the literature, Dubois and Prade [8, 9, 6] propose a qualitative decision theory based on possibility theory. There are certain facts that make their proposal comparable to our study. First, there is a close relation between possibility theory and Spohn's epistemic belief theory as pointed out in [7]. The possibility of a proposition $\pi(x)$ is related to the degree of disbelief by the relation $\pi(x) = \exp(-\delta(x))$, and vice versa. Second, Dubois and Prade base their proposal on von Neumann and Morgenstern's axioms as we do in this study.

However, there are several important differences. First, the setting in [8] is to qualitatively compare belief states given a fixed act where as the setting in our study is to compare acts given a fixed state of belief. The setting of [8] sometimes leads to a confusion about the intuition behind the preference relation on belief states. For example, Dubois and Prade's preference relation $\succeq_{DP}$ on the set of possibility distributions, by definition, is supposed to reflect comparison on utility. However, their Axiom 3: If $\pi \leq \pi'$ then $\pi \succeq \pi'$ ("precision is safer") is imposed just by informational consideration. This axiom is contradicts our axiom 6 ("qualitative monotonicity"). When proposing Axiom 6, we have stated the rationale for adopting it. Let us consider an example. There are two situations, $x$ denotes a loss of \$1,000 and $y$ denotes a gain of \$1,000,000. Dubois-Prade's Axiom 3 would suggest that the loss of \$1,000 with certainty has no less utility than a lottery in which a loss of \$1,000 and a gain of \$1,000,000 are equally possible. However, according to our Axiom 6, the latter lottery is preferred to the former.

Another difference between Dubois-Prade's and our proposals is in how compound lotteries are handled as summarized in their Axiom 5 and our Axiom 2. In our proposal, as in von Neumann-Morgenstern theory, the notion of independence between betting stages is exploited to derive the rule for reducing multi-stage lotteries. In Dubois-Prade's Axiom 5, just conservative reasoning is invoked. We feel that incorporating independence information in the rule for manipulating compound lotteries, like in von Neumann-Morgenstern's work, makes a decision theory more realistic and practical than ignoring it. Although we are aware of problems and difficulties in justifying independence when all one has is information about irrelevance or lack of interaction, in practice people do perceive those notions interchangeably. In some situations, people may behave very cautiously, but they often rationally engage in risky business if the risk is reasonable for them.

In [6], for a possibilistic lottery that is defined in a way similar to our lottery construction, there are two kinds of utilities called the "pessimistic" and "optimistic" utilities that are obtained by using two different sets of rules. Obviously, optimal decision depends on which utility is employed. In other words, there is information about the meta-preference provided by users that is not covered within the formal systems proposed by the authors. In contrast, our utility theory doesn't make any assumption about the risk attitude of the decision maker as in the case of von Neumann and Morgenstern's theory. This feature allows us to avoid the ambiguity a user faces when she wants to use possibilistic utility.

However, it is important to note that in simple situations like the one in Savage's omelette example, a decision maker using a pessimistic utility function can be modeled as a user who follows our qualitative utility theory.

An interesting line of work in qualitative decision have been pursued by Brafman and Tennenholtz [3, 4]. The authors adopt an axiomatic approach argued by Savage [17] and characterize conditions under which an agent can be said as using *maximin, minimax regret, competitive ratios and maximax* decision criteria. They show that these very different criteria are equivalent in terms of representation power. In particular, the representation theorem for maximin rule says that if an agent's preference satisfies a property similar to Savage's *sure thing principle* and a *transitive-like* property then the agent's decision can be modeled by maximin rule. Purely qualitative rule such as maximin is justified because in their setting the consideration of chance or likelihood of possible worlds is ignored. In our setting, the notion of beliefs of possible worlds enters explicitly in decision making. It can be easily shown that preference based on the notion of expected qualitative utility we develop here cannot be modeled by a simple maximin rule.

Qualitative decision based on Spohn's calculus has also been studied by Pearl and his associates at UCLA [10, 16, 11, 22, 5, 2] and Wilson [23]. They show that a disbelief degree (ranking) can be viewed as order-of-magnitude probabilities. A similar idea can be traced back to Adams [1] (see [15] for a discussion). These authors successfully use the relationship between kappa rankings and probability to solve problems in non-monotonic reasoning, for example, defining probabilistic semantics for default rules and conditional ought statements. They provide various set of decision making rules justified by semantics of non-standard probability. However, none of these studies approach the problem from an axiomatic point of view as we do in this paper. We conjecture that most of these results can be justified by using our qualitative utility function and some assumptions of the nature of the utility function.

## 6 Summary and Conclusions

Our goal in this paper is to propose a utility theory for Spohnian lotteries in which the uncertainty of winning different prizes is expressed by epistemic beliefs. The utility function obtained is qualitative in the sense that a discrete scale is used. Also, in the formula of expected utility, minimization and addition operations are used in place of addition and multiplication in the formula for quantitative expected utility. This seems to make sense for lotteries in which uncertainty is characterized qualitatively by disbelief values. Methodologically, we adapt the construction of a linear utility function for the case of probabilistic lotteries to the case of Spohnian lotteries. We show that preference among Spohnian lotteries that is required to satisfy some plausible axioms can be represented by an analogous qualitative linear utility function.

Decision making based on qualitative expected utility is somewhere in between purely qualitative rules such as maximin, minimax regret or maximax on one hand, and the purely quantitative rule of maximizing von Neumann-Morgenstern's expected utility on the other hand. Unlike, for example, the maximin rule which focuses on the worst possible outcome, our qualitative utility theory incorporates the epistemic beliefs about realization of all possible outcomes. However, qualitative utility may be viewed as order-of-magnitude approximation of quantitative utility. We think that the position in the middle ground between the two camps is a good one. That avoids criticisms both camps make toward the other. For quantitative utility, a common critique is that it often demands more than realistically available assessments of uncertainty and preferences for standard lotteries. For decision rules such as maximin, the critique is that it is too conservative and often ignores information even when available. It is easy to find examples in which maximin rule leads to unrealistic choices. Although we do not claim that the earthquake example is a realistic one, it helps to illustrate our point. In order to calculate vNM utility, we are suppose to have a probability distribution of earthquakes (despite the fact that earthquake specialists say that these are hard to come by) and we need to assess precise utilities for each of the earthquake intensities. On the other hand, if the maximin rule were used, people would base their decisions on the effects of intensity 12 earthquake which is not realistic. Our qualitative utility theory demands assessments of uncertainty in terms of epistemic beliefs and assessments of qualitative utilities for each of the earthquake intensities, a reasonable middle ground between the two extremes.

### Acknowledgements

We thank anonymous referees for their comments and suggestions to improve the presentation of this paper.